\def\year{2020}\relax
\documentclass[letterpaper]{article} % DO NOT CHANGE THIS
\usepackage{aaai20}  % DO NOT CHANGE THIS
\usepackage{times}  % DO NOT CHANGE THIS
\usepackage{helvet} % DO NOT CHANGE THIS
\usepackage{courier}  % DO NOT CHANGE THIS
\usepackage[hyphens]{url}  % DO NOT CHANGE THIS
\usepackage{graphicx} % DO NOT CHANGE THIS
\usepackage{graphicx}  % DO NOT CHANGE THIS
\usepackage{tikz-dependency}
\usepackage{tikz}
\usepackage{tikz-qtree}
\newcommand{\citet}[1]{\citeauthor{#1} \shortcite{#1}}

\title{Discontinuous Constituent Parsing with Pointer Networks}
\author{Daniel Fern\'{a}ndez-Gonz\'{a}lez \ and \ Carlos G\'{o}mez-Rodr\'{i}guez\\
Universidade da Coru\~{n}a, CITIC \\
FASTPARSE Lab, LyS Group \\
Depto. de Ciencias de la Computaci\'{o}n y Tecnolog\'{i}as de la Informaci\'{o}n \\
Elvi\~{n}a, 15071 A Coru\~{n}a, Spain \\
d.fgonzalez@udc.es, carlos.gomez@udc.es
}

\begin{document}
\maketitle
\begin{abstract}
One of the most complex syntactic representations 
used in computational linguistics and NLP
are discontinuous constituent trees, crucial for representing all grammatical phenomena of languages such as German. Recent advances in dependency parsing have shown that Pointer Networks excel in efficiently 
parsing
syntactic relations between words in a sentence. This kind of sequence-to-sequence models achieve outstanding accuracies in building non-projective dependency trees, but its potential has not been proved yet on a more difficult task. We propose a novel neural network architecture that, by means of Pointer Networks, is able to generate the most accurate discontinuous constituent representations to date, even without the need of Part-of-Speech tagging information. To do so, we internally model discontinuous constituent structures as augmented non-projective 
dependency structures.
The proposed approach achieves state-of-the-art results on 
the two widely-used NEGRA and TIGER benchmarks, 
outperforming previous work by a wide margin.
\end{abstract}

\section{Introduction}
Syntactic representations are increasingly being demanded by a wide range of artificial intelligence applications that process and understand natural language text or speech. This encourages the Natural Language Processing (NLP) community to develop more accurate and efficient parsers that are able to represent all complex grammatical phenomena present in human languages. 

\begin{figure*}[t]
\centering
\hspace{0.6cm}\includegraphics[width=0.32\textwidth]{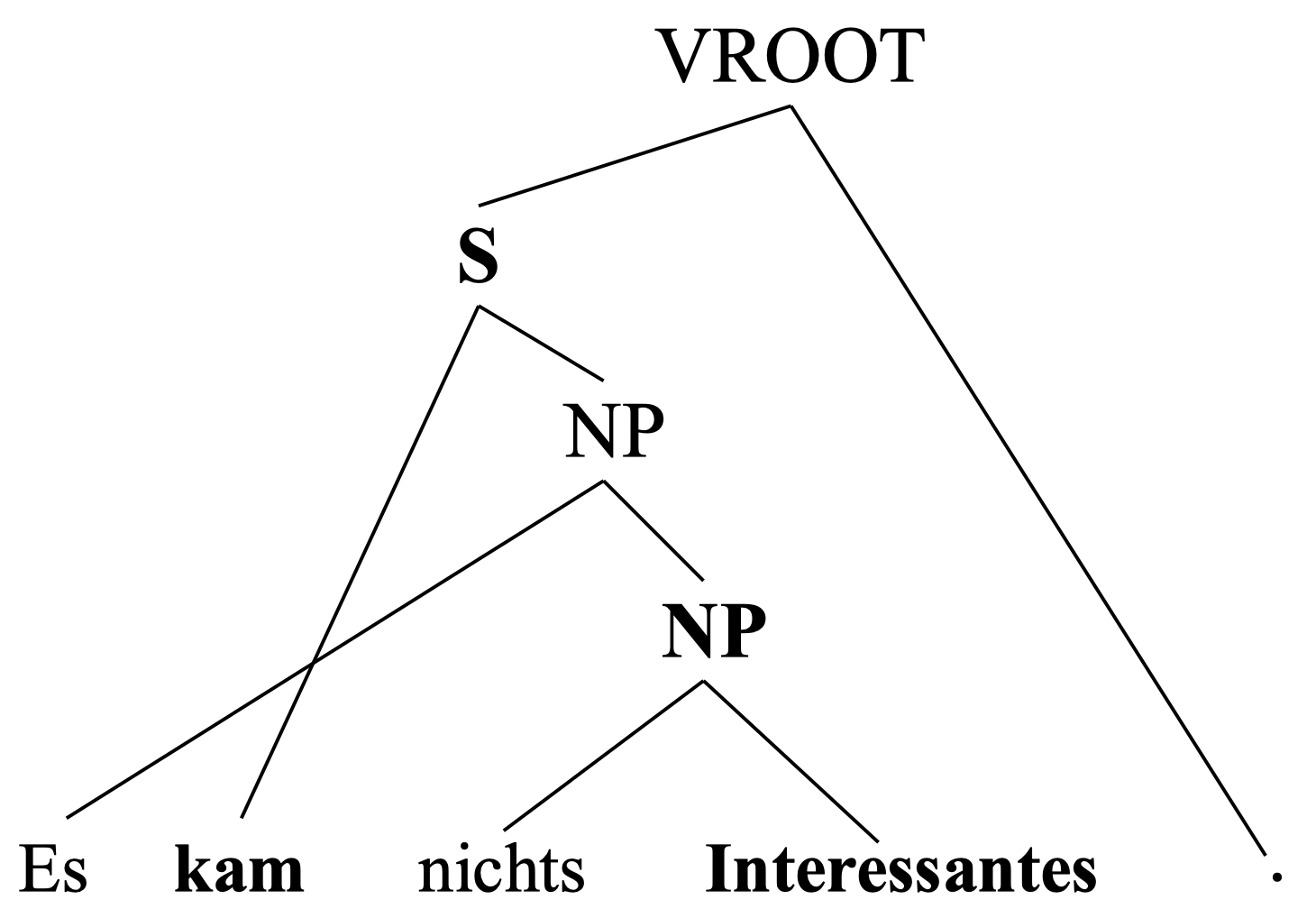}
\qquad\qquad
\hspace{0.4cm}\includegraphics[width=0.35\textwidth]{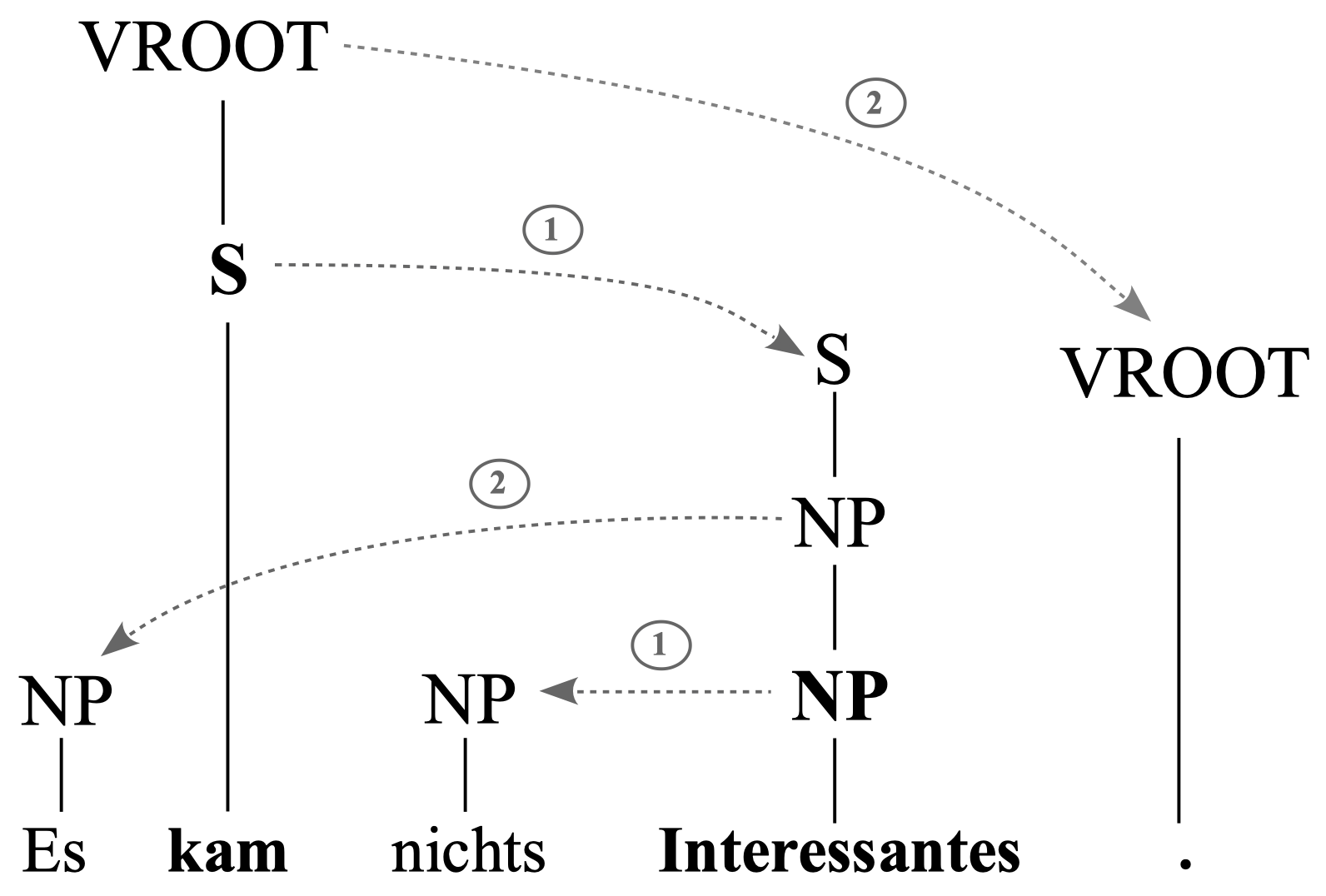}\\
\small{a) Discontinuous constituent tree.} \qquad\qquad\qquad \hspace{2.6cm}\small{b) Spines interaction.}\\
\vspace{0.3cm}
\begin{dependency}[theme = simple]
\begin{deptext}[column sep=1.3em]
Es \& kam \& nichts \& Interessantes \& . \\
\end{deptext}
\depedge[edge unit distance=2ex]{4}{1}{app}
\depedge{4}{3}{det}
\depedge[edge unit distance=4.5ex]{2}{4}{subj}
\depedge[edge unit distance=4ex]{2}{5}{punct}
\end{dependency}
\qquad\qquad
\begin{dependency}
\begin{deptext}[column sep=1.3em]
Es$_1$ \& kam$_2$ \& nichts$_3$ \& Interessantes$_4$ \& .$_5$ \\
\end{deptext}
\depedge[edge unit distance=2ex]{4}{1}{NP\#2}
\depedge{4}{3}{NP\#1}
\depedge[edge unit distance=4.5ex]{2}{4}{S\#1}
\depedge[edge unit distance=4ex]{2}{5}{VROOT\#2}
\end{dependency}

\small{c) Labelled dependency tree.} \qquad\qquad\qquad \hspace{2.4cm}d) \small{Augmented dependency tree.}\\
\caption{Augmented dependency representation of a sentence taken from NEGRA corpus in discontinuous constituent format, as well as the spine interaction required for its reconstruction. Head nodes of each constituent are marked in bold.}
\label{fig:trees}
\end{figure*}

One of the most widely-used syntactic formalism is a \emph{constituent} (or \textit{phrase-structure}) representation.
This describes the syntax of a sentence in terms of constituents or phrases and the hierarchical order between them, as shown in the constituent tree in Figure~\ref{fig:trees}(a). As a simpler alternative, a dependency representation is also available. This straightforwardly connects each word of a sentence as a dependent of another, which acts as its head word. The resulting structure is a dependency tree like the one depicted in Figure~\ref{fig:trees}(c).

Among constituent trees we can find the most informative syntactic representation currently available: \textit{discontinuous} constituent trees. Apart from providing phrase-structure information, they extend regular or \textit{continuous} constituent trees by allowing the representation of crossing branches and constituents with gaps in the middle, crucial for describing grammatical structures left out in the standard phrase-structure formalism, such as the free-word-order phenomena that can be found in languages such as German 
(see \cite{Muller2004} and references therein).
%\cite{Muller2004,maier:lichte:11}. 

 Unlike for regular constituent trees
 \cite{klein03}, \textit{context-free grammars} are not enough for deriving discontinuous structures and representing complex linguistic phenomena and, therefore, more expressive formalisms, such as \textit{Linear Context-Free Rewriting Systems} (LCFRS) \cite{LCFRS}, are required. However, due to the high complexity behind the generation of discontinuous phrase-structure representations, parsers based on probabilistic LCFRS, such as 
 \cite{kallmeyer2010}, are not practical in terms of accuracy and speed.
 In the last decade, alternatives to tackle discontinuous constituent parsing with complex grammar-based approaches were presented.

On the one hand,
one line of research proposes
to extend transition-based parsers, which have been successfully applied for efficiently building continuous constituent trees \cite{Zhu13,Dyer2016}, 
with transitions and data structures that are able to deal with discontinuity 
 \cite{maier2015,stanojevic2017,coavoux2019a,coavoux2019b}. This research branch currently yields the state of the art in discontinuous phrase-structure parsing. 

In parallel, another stream led by \cite{Hall2008,versley2014,reduction,corro2017} proposes to represent discontinuous formalisms as dependencies with augmented information.
This makes it possible to use efficient dependency parsers to perform the analysis and, after a recovery process, return a well-formed constituent tree. 
While they offer a similar efficiency as 
transition-based constituent parsers,
they are currently slightly behind   
them
in terms of accuracy. The main drawback of this approach is the large amount of complex labels necessary to encode the wide variety of syntactic structures into labelled dependency arcs, growing unbounded with the treebank size.
This forces some authors, as \cite{reduction,corro2017}, to use an extra module that, at a post-processing step after the parsing, labels each dependency to increase final accuracy. As a result, parsing and labelling tasks are learned and performed in a two-stage procedure.
However, these two tasks
might benefit from each other's information during training, as constituent information is encoded in both components (arc and label) 
so from an ideal standpoint, given good enough machine learning models, they should not be considered separately.

To address this problem, we propose a neural network architecture that, using internal augmented dependency representations based on \cite{reduction} and following a multitask learning approach \cite{multitask}, can directly produce accurate discontinuous constituent trees in one step. Both tasks (creating dependencies and labeling them) are trained in parallel and using a shared representation, so that the whole model can jointly learn to predict the correct augmented labels that will finally produce well-formed phrase-structure trees. 

In addition, our model relies on \textit{Pointer Networks} \cite{Vinyals15} for creating dependency arcs between words of an input sentence. This kind of neural networks, obtained by modifying standard sequence-to-sequence models, yields remarkable accuracies in dependency parsing. In particular, they use attention \cite{Bahdanau2014} as a pointer to select positions from the input sequence, which can be easily adapted to connect words from an input sentence, as recently shown by \cite{Ma18,L2RPointer}. We adapt this state-of-the-art neural model to a harder task, also obtaining an outstanding performance. To the best of our knowledge, this is the first attempt that follows a sequence-to-sequence paradigm to perform discontinuous constituent parsing, without the need of any intermediate data structures to create partial trees as required by transition-based models previously cited. 

To deal with the large amount of labels, we propose to use the biaffine classifier introduced by \cite{DozatM17} that has shown a remarkable performance even in NLP taks with a more complex label scheme, such as Semantic Dependency Parsing \cite{dozat-manning-2018}. This shares the same representation as Pointer Networks and is jointly trained to produce well-formed discontinuous constituent trees in a single step.

The resulting neural model\footnote{Available at \url{https://github.com/danifg/DiscoPointer}} produces the most accurate discontinuous constituent representations reported so far. We conduct experiments on both widely-known NEGRA \cite{Skut1997} and TIGER \cite{brants02} treebanks, which contain a large number of German sentences syntactically annotated by constituent trees with a high degree of discontinuity: the prevalence of discontinuous constituent structures is over 25\% in both treebanks \cite{maier:lichte:11}. In both benchmarks, our approach achieves accuracies beyond 85\% F-score (even without Part-of-Speech (POS) tagging information), surpassing the current state of the art by a wide margin without the need of orthogonal techniques such as re-ranking or semi-supervision.

\section{Modelling Discontinuous Constituent Trees}
\subsection{Preliminaries}
Let $w_1, w_2, \dots, w_n$ be a sentence, where $w_i$ denotes a word in the $i$th position. A constituent tree is 
a tree with 
the $n$ words of the sentence as leaf nodes, and phrases (or constituents) as internal nodes. Each constituent can be represented as a tuple $(X, \mathcal{C}, w_h)$, where $X$ is the non-terminal symbol, $\mathcal{C}$ is the yield, i.e. the set of words $w_i$ included in its span, and $w_h$ is the word in $\mathcal{C}$ that acts as head. A language-specific handwritten set of rules is used to select the head word. For instance, the head word of the constituent $\mathit{VROOT}$ in Figure~\ref{fig:trees}(a) is the word $\mathit{kam}$ in constituent $S$.

Given a tree, if the yield ($\mathcal{C}$) of each of its constituents is a continuous substring of the sentence,
then
we say that the tree is \textit{continuous}. Otherwise, the tree is classified as \textit{discontinuous}, meaning that we can find at least one constituent with a span that 
is interrupted by one or more gaps between its words.
For instance, in Figure~\ref{fig:trees}(a), the span of constituent $(\mathit{NP}, \{\mathit{Es}, \mathit{nichts}, \mathit{Interessantes}\}, \mathit{Interessantes})$ 
is interrupted by
the word $kam$ from a different constituent, generating crossing branches.
Finally, if there are no constituents with only one child node, we call it a \textit{unaryless} constituent tree.

On the other hand, a dependency tree is a directed tree spanning all the words $w_i$ with $i$ ranging from 1 to $n$. Each dependency arc can be represented as $(w_h,w_d,l)$, with $w_h$ being the head word of the dependent word $w_d$ (with $h$, $d \in [1,n]$) and $l$ being the dependency label indicating the syntactic role played by $w_d$. If, for every dependency arc $(w_h,w_d,l)$, there is a directed path from $w_h$ to all words $w_i$ between words $w_h$ and $w_d$, we classify it as a \textit{projective} dependency tree. If not, it is described as \textit{non-projective}, resulting in a tree with crossing arcs as the one depicted in Figure~\ref{fig:trees}(c).

It can be noticed from these definitions that, in order to represent the same syntactic phenomenon as described in a discontinuous constituent tree, we will need to use a non-projective dependency structure in order to handle discontinuities. However, a constituent tree is able to provide information that cannot be represented in a regular dependency tree \cite{Kahane2015}.

\subsection{Constituents as Augmented Dependencies}

Following \cite{reduction}, we
decompose a discontinuous unariless phrase-structure tree into a set of non-projective dependency arcs with enriched information. This allows us to use a non-projective dependency parser to efficiently perform discontinuous phrase-structure parsing. 

A constituent tree with $m$ words can be decomposed into a set of $m$ spines \cite{carreras08}, one per word, as shown in Figure~\ref{fig:trees}(b) for the discontinuous tree in Figure~\ref{fig:trees}(a). These spines and their interaction to finally build a constituent tree can be represented in an augmented dependency tree as described by \cite{reduction}. More in detail, they propose to encode each constituent into dependencies: 
for each constituent $(X,\mathcal{C},w_h)$, each child node, that can be a word $w_d$ or a constituent $(Y,\mathcal{G},w_d)$ with $w_d \neq w_h$, is encoded into an unlabelled dependency arc with the form $(w_h, w_d)$. Basically, each non-head child node is attached to the head word and, if the involved child node is a constituent, its own head word is used to build the dependency arc.

Additionally, to represent the original phrase structure it is also necessary to save 
some vital information into arc labels: the non-terminal symbol $X$ plus an index that indicates the order in which spines are attached. This index will be crucial in those cases where more than one constituent share the same head word (and, therefore, the same head spine), but they are at a different level in the original tree and, therefore, should be created in a different hierarchical order. The final dependency arcs will have the form $(w_h,w_i,X\#p)$, where $p$ is the attachment order. For instance, constituent $(\mathit{NP}, \{\mathit{nichts}, \mathit{Interessantes}\}, \mathit{Interessantes})$ in Figure~\ref{fig:trees}(a), is encoded as the augmented dependency arc $(\mathit{Interessantes}, \mathit{nichts}, \mathit{NP\#1})$ in Figure~\ref{fig:trees}(d), and constituent $(\mathit{NP}, \{\mathit{Es}, \mathit{nichts}, \mathit{Interessantes}\},$ $ \mathit{Interessantes})$ is represented with $(\mathit{Interessantes}, \mathit{Es}, \mathit{NP\#2})$. Both constituents share the same head spine anchored to word $\mathit{Interessantes}$, but they are attached in a different level.

To recover the original discontinuous constituent tree, we just have to follow the steps provided by the augmented dependencies from a spine-based perspective. Dependencies will indicate which spines are involved in the attachment and the direction of the spine interaction (head and dependent roles);
and, on the other hand, labels will describe in what order dependent spines are attached to the head spine, as well as the non-terminal symbol of the resulting constituent. For instance, the augmented dependency tree in Figure~\ref{fig:trees}(d) describes the spine interaction depicted in Figure~\ref{fig:trees}(b). In the example, we can see that the dependent spines anchored to words \textit{Es} and \textit{nichts} are attached in different order and level from the head spine lexicalized by the word \textit{Interessantes}. 
Without the attachment order, all dependent spines would be attached at the same level and the resulting constituent would be a different structure: in the example, a single flat NP phrase with the three words as its child nodes.

It is also worth mentioning that the above described dependency-based representation is not able to encode unary constituents and, as a consequence, this information will be lost after the constituent tree is recovered. However, as stated by \cite{reduction}, unary nodes are very uncommon in discontinuous representations: for instance, the NEGRA treebank has no unaries at all and, for the TIGER dataset, the fraction of unaries is around 1\%. Therefore, we do not perform unary recovery in a postprocessing step and simply apply a non-projective dependency parser on the resulting augmented trees.

\section{Neural Network Architecture}
We use a multi-task learning strategy to train in parallel both the Pointer Network, in charge of connecting words, and the labeler, a multiclass classifier in charge of tagging each dependency arc with the augmented information, to finally build a well-formed discontinuous constituent tree. In the next subsections, we describe the proposed pointer-network-based architecture for labelled dependency parsing.

\subsection{Pointer Networks}
\cite{Vinyals15} introduced a novel neural architecture, called \textit{Pointer Network}, that can be seen as a variation of standard sequence-to-sequence models. The proposed neural network can learn the conditional probability of an output sequence of discrete numbers that correspond to positions from the input sequence, with input length being variable. They propose to use a mechanism of neural attention \cite{Bahdanau2014} to select positions from the input, without requiring a fixed size of the output dictionary. Thanks to that, Pointer Networks are suitable for addressing those problems where the target classes considered at each step are variable and depend on the length of the input sequence. 

Unlabelled dependency parsing is one of the tasks where Pointer Networks can be easily applied: the input is a (variable-length) sequence of words from a sentence and the output is a sequence of numbers that correspond to the positions of the assigned head words.
In \cite{L2RPointer}, we can find a transition-based perspective of this idea, but it can be also defined as a purely sequence-to-sequence approach, as, unlike in classic transition-based dependency parsers \cite{nivre03iwpt}, neither transitions nor data structures are required.

We follow the convention of representing scalars in lowercase italics ($i$), vectors in lowercase bold ($\mathbf{v}$), matrices in uppercase italics ($M$), and higher order tensors in uppercase bold ($\mathbf{T}$).

\begin{figure*}[t]
\centering
\includegraphics[width=0.8\textwidth]{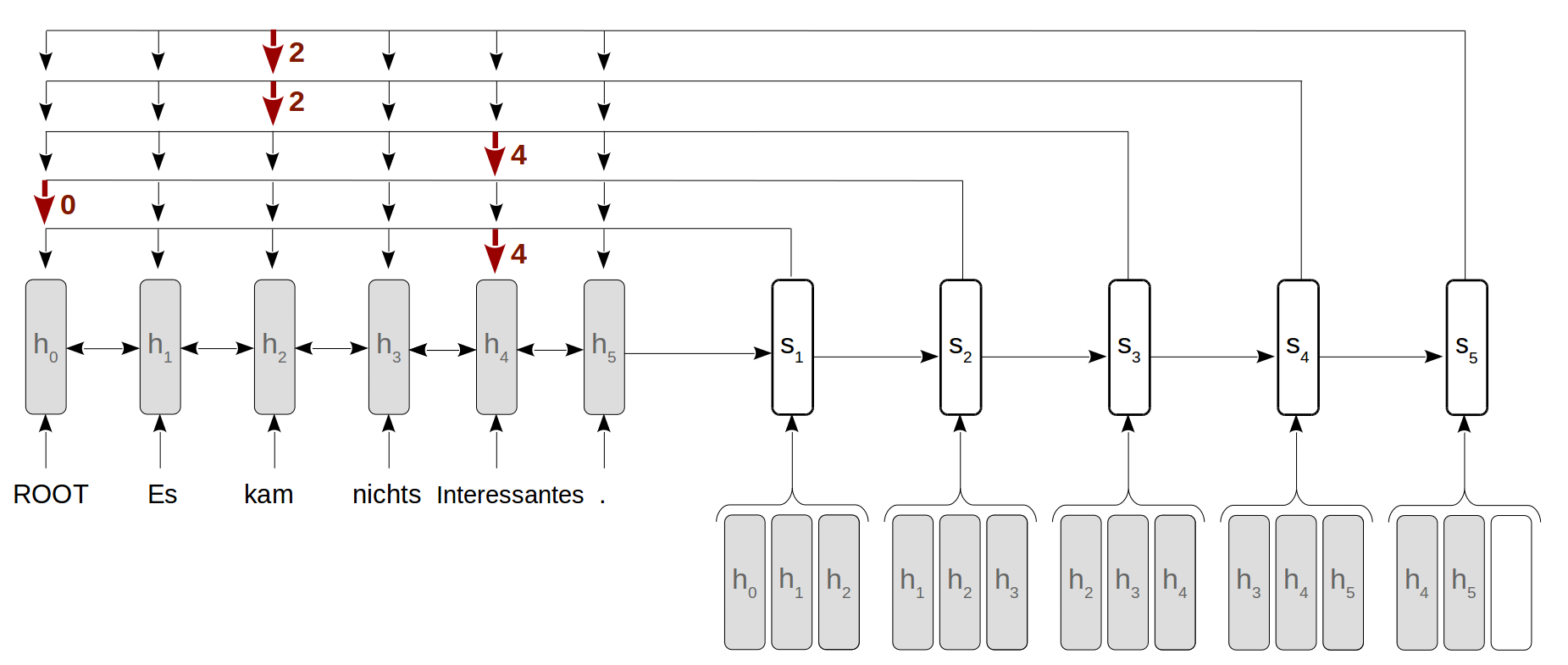}
\caption{Pointer Network architecture and decoding steps to output the dependency tree in Figure~\ref{fig:trees}(d).}
\label{fig:network}
\end{figure*}

\subsubsection{Encoder}
Let $w_1, w_2, \dots ,w_n$ be an input sentence. The encoder of our network relies on a BiLSTM-CNN architecture, developed by \cite{Ma2016}, to encode each word $w_i$
into an \textit{encoder hidden state} $\mathbf{h}_i$. 

Convolutional Neural Networks (CNN) \cite{LeCun1989} are used for extracting character-level representations of words. In particular, for each word $w_i$, the CNN receives a concatenation of embedding vectors of each character as input and produces a character-level representation $\mathbf{e}^c_i$. Then, each word $w_i$ is represented by the concatenation of three embeddings: character-level word embedding $\mathbf{e}^c_i$, pretrained word embedding $\mathbf{e}^w_i$ and randomly initialized part-of-speech (POS) tag embedding $\mathbf{e}^p_i$ ($\oplus$ stands for concatenation):
$$\mathbf{x}_i = \mathbf{e}^c_i \oplus \mathbf{e}^w_i \oplus \mathbf{e}^p_i$$
In order to attach every word from the sentence to another, we need to add a dummy root node $w_0$ (represented with a special vector $\mathbf{x}_0$) at the beginning of the sentence. This will be linked as the head of the real syntactic root word.

After that, the resulting embedding representation $\mathbf{x}_i$ of each word $w_i$ is fed one-by-one into a Bi-directional Long Short-Term Memory (BiLSTM) \cite{Graves2005} that captures context information in both directions and generates a vector representation $\mathbf{h}_i$:
$$ \mathbf{h}_i = \mathbf{h}_{li}\oplus\mathbf{h}_{ri} = \mathbf{BiLSTM}(\mathbf{x}_i)$$

\subsubsection{Decoder}
A unidirectional LSTM \cite{Hochreiter1997} is used as a decoder that, at each time step, uses the encoder hidden state $\mathbf{h}_i$ of the word $w_i$, currently being processed, as input and outputs a \textit{decoder hidden state} $\mathbf{s}_t$. To add extra information to the input of the LSTM, we follow \cite{L2RPointer} and add hidden states of previous ($\mathbf{h}_{i-1}$) and next ($\mathbf{h}_{i+1}$) words. As proposed by \cite{Ma18}, we use element-wise sum of the three hidden states, instead of concatenating them, in order to not increase the dimension of the resulting input vector $\mathbf{r}_i$:
$$\mathbf{r}_i = \mathbf{h}_{i-1} + \mathbf{h}_{i} + \mathbf{h}_{i+1}$$
$$ \mathbf{s}_t = \mathbf{LSTM}(\mathbf{r}_i)$$
Once $\mathbf{s}_t$ is generated from $\mathbf{r}_i$, the \textit{attention vector} $\mathbf{a}^t$ needs to be computed as follows:
$$\mathbf{v}^t_j = \mathbf{score}(\mathbf{s}_t, \mathbf{h}_j)$$
$$\mathbf{a}^t = \mathbf{softmax}(\mathbf{v}^t)$$
where an \textit{attention scoring function} ($\mathbf{score}()$) is used to compute scores between $\mathbf{s}_t$ (the decoder representation of the word $w_i$ currently being processed) and each encoder hidden state $\mathbf{h}_j$ from the input. Then, a softmax is applied on the resulting score vector $\mathbf{v}^t$ to compute a probability distribution over the input. The generated attention vector $\mathbf{a}^t$ is used as a pointer to select the highest-scoring position $j$ from the input. The pointed word $w_j$ is attached as head of word $w_i$, currently being processed, building an unlabeled dependency arc ($w_j$, $w_i$). In case the arc ($w_j$, $w_i$) generates a cycle in the already-built dependency graph, 
we select the next highest-scoring position from the input pointed by $\mathbf{a}^t$.

The decoding process starts in word $w_1$ and it is repeated word-by-word from left to right until the last one ($w_n$) is attached, requiring just $n$ steps to fully parse the input sentence. Therefore, since cycles can be checked in amortized linear time ($O(n)$) and, at each step, the attention vector $\mathbf{a}^t$ must be computed over the input, the overall runtime complexity is $O(n^2)$.  Figure~\ref{fig:network} depicts the neural architecture and the decoding procedure for the dependency structure in Figure~\ref{fig:trees}(d).

\subsubsection{Attention Mechanism}
As attention score function, we follow \cite{Ma18} and adopt the biaffine attention mechanism, developed by \cite{DozatM17}, to compute score vector $\mathbf{v}^t$:
$$\mathbf{v}^t_j = \mathbf{s}^T_t W \mathbf{h}_j + \mathbf{U}^T\mathbf{s}_t + \mathbf{V}^T\mathbf{h}_j + \mathbf{b}$$
where parameters $W$ is the weight matrix of the bi-linear term, $\mathbf{U}$ and $\mathbf{V}$ are the weight tensors of the linear terms and $\mathbf{b}$ is the bias vector. Basically, the bi-linear term evaluates the score of assigning the word $w_i$ (represented by $\mathbf{s}_t$) to the head word $w_j$ from the input, and the two linear terms compute the scores of both words considered independently. 

Finally, in order to reduce the dimensionality and overfitting of the neural model, a multilayer perceptron (MLP) is used to transform the hidden representations $\mathbf{h}_j$ and $\mathbf{s}_t$ before the attention score function is applied, as discussed by \cite{DozatM17}.  

\subsection{Biaffine labeler}
 We follow \cite{DozatM17} and implement the labeler as a biaffine multi-class classifier. Unlike them, we use the previously described BiLSTM-CNN architecture as encoder.
 
 \cite{DozatM17} relies on a biaffine attention mechanism as scoring function for computing the score of assigning a label $l$ on a predicted arc between the dependent word $w_i$ and the head word $w_j$, assigning to that arc the highest-scoring label. 
 In particular, the encoder hidden state $\mathbf{h}_j$ and the decoder hidden state $\mathbf{s}_t$ (representing  $w_j$ and $w_i$, respectively)
 are used as input to obtain the score $\mathbf{s}^l_{tj}$ for each label $l$ as shown in the following equation:
$$\mathbf{s}^l_{tj} = \mathbf{s}^T_t W_l \mathbf{h}_j + \mathbf{U}_l^T\mathbf{s}_t + \mathbf{V}_l^T\mathbf{h}_j + \mathbf{b}_l$$
where a distinct weight matrix $W_l$, weight tensors $\mathbf{U}_l$ and $\mathbf{V}_l$ and bias $\mathbf{b}_l$ are used for each label $l$, where $l \in \{1, 2, \dots , L\}$ and $L$ is the number of labels. The first term on the right side of the equation represents the score of assigning the label $l$ to the arc between dependent $w_i$ and head $w_j$; and the second and third terms express the score of the label $l$  when the dependent and head are considered independently. Please note that a MLP is also used to transform hidden state vectors $\mathbf{h}_j$ and $\mathbf{s}_t$, before feeding the biaffine classifier, in order to filter not relevant information and, as a consequence, reduce model dimensionality \cite{DozatM17}.

\section{Multitask Learning}
Our 
goal is to provide a fully-parsed non-projective dependency tree in a single stage, while the parsing and the labelling should be individually undertaken due to the large amount of labels resulting from the formalism by \cite{reduction}. We follow a multitask learning strategy \cite{multitask} to achieve that: a single neural architecture is trained for more than one task, sharing a common representation and benefiting from each other.

In particular, the labeler shares the same encoder as the parser, providing a common encoder hidden state representation for both components.  As stated by \cite{Kiperwasser2016}, training the BiLSTM-CNN encoder to correctly predict dependency labels significantly improves unlabelled parsing accuracy and vice versa. This is specially crucial in our approach where a constituent structure is jointly encoded in dependency arcs and labels, and a wrong label can lead to a completely different phrase-structure tree after the recovering.

In addition, the labeler and the parser are simultaneously trained by optimizing the sum of their objectives. On the one hand, a dependency tree $y$ for an input sentence of length $n$ is decomposed into a set of $n$ directed arcs $a_1, \dots , a_{n}$, where each arc $a_i$ is represented by the head word $w_h$ and the dependent word $w_i$ in position $i$ in the sentence. Therefore, to train the parser, we factorize the conditional probability $P_\theta (y|x)$ of a dependency tree $y$ for a sentence $x$ into a set of head-dependent pairs ($w_h$, $w_i$) as follows:
$$P_\theta (y|x) = \prod_{i=1}^{n} P_\theta (a_i | a_{<i}, x)
= \prod_{i=1}^{n} P_\theta (w_h | w_i,a_{<i},x)$$
 Then, the unlabelled parsing model is trained by minimizing the negative log likelihood of choosing the correct head word $w_h$ for the word $w_i$ currently being processed, given the previous predicted arcs $a_{<i}$. This is implemented as cross-entropy loss:
$$\mathcal{L}_{arc} = -log P_\theta (w_h | w_i,a_{<i},x)$$

On the other hand, the biaffine labeler is trained by minimizing the negative log probability of assigning the correct label $l$, given a dependency arc 
with
head word $w_h$ and dependent word $w_i$. 
A cross entropy loss $\mathcal{L}_{label}$ for dependency label prediction is computed when training:
$$\mathcal{L}_{label} = -logP_\theta (l|w_h, w_i)$$

Finally, we train a joint model by summing the losses $\mathcal{L}_{arc}$ and $\mathcal{L}_{label}$ prior to computing the gradients. In that way, model parameters are learned to minimize the sum of the cross-entropy loss objectives over the whole corpus.

\section{Experiments}
\subsection{Data}
We test our new approach on two widely-used discontinuous German treebanks: NEGRA \cite{Skut1997} and TIGER \cite{brants02}. For the latter, we use the split provided in the SPMRL14 shared task \cite{SPMRL}, and, for NEGRA, we follow the standard splits \cite{dubey2003}. While in TIGER we use the predicted POS tags provided in the shared task, we run TurboTagger \cite{turboparser} for predicting POS tags for NEGRA. Finally, for both treebanks, we apply the head-rule sets by \cite{Rehbein2009} to identify head words in constituent structures and build the equivalent augmented dependency trees. Following standard practice, we use discodop\footnote{\url{https://github.com/andreasvc/disco-dop}} \cite{Cranenburgh2016} with the configuration file \texttt{proper.prm} for evaluation. This ignores punctuation and root symbols.

\subsection{Settings}
We use the Adam optimizer \cite{Adam} and follow \cite{Ma18,DozatM17} for parameter optimization and hyper-parameter selection. These are detailed in Table~\ref{tab:hyper}. All embeddings are fine-tuned during training and, for initializing word vectors, we use the pre-trained structured-skipgram embeddings developed by \cite{Ling2015}. Due to random initializations, we report average accuracy over 5 repetitions for each experiment tested. In addition, during training, the model with the highest Labelled Attachment Score (LAS) on the augmented dependency version of the development set is chosen. Finally, we use 10-beam-search decoding for all experiments.  

\begin{table}
\begin{footnotesize}
\small
\centering
\begin{tabular}{@{\hskip 0pt}lc@{\hskip 0pt}}
\hline
\textbf{Architecture hyper-parameters} & \\
\hline
CNN window size & 3 \\
CNN number of filters & 50 \\
BiLSTM encoder layers & 3 \\
BiLSTM encoder size & 512 \\
LSTM decoder layers & 1 \\ 
LSTM decoder size & 512 \\
POS tag/word/character embedding dimension & 100 \\
LSTM layers/embeddings dropout & 0.33 \\
MLP layers & 1 \\
MLP activation function & ELU \\
Arc MLP size & 512 \\ 
Label MLP size & 128 \\
\hline
\textbf{Adam optimizer hyper-parameters} &\\
\hline
Initial learning rate & 0.001 \\
$\beta_1$, $\beta_2$ & 0.9 \\
Decay rate & 0.75 \\
Gradient clipping & 5.0 \\
\hline
\multicolumn{1}{c}{}\\
\end{tabular}
\centering
\setlength{\abovecaptionskip}{4pt}
\caption{Model hyper-parameters.}
\label{tab:hyper}
\end{footnotesize}
\end{table}

\subsection{Results}
In Table~\ref{tab:results}, we report the overall F-score on labeled constituents and a specific F-score measured only on discontinuous constituents (Disc. F1) of current state-of-the-art parsers in comparison to our neural model. 

As seen in the table, the proposed neural model is able to outperform all existing systems by a wide margin in both predicted- and gold-POS-tagging scenarios. While gold POS tags prove to be beneficial for the parsing process, using predicted ones yields the same performance as not providing POS-tagging information at all. Finally, we can also conclude from these results that, even without neither pre-trained word embeddings nor POS tags, our approach still surpasses the current state of the art by a wide margin.

\begin{table*}
\small
\centering
\begin{tabular}{@{\hskip 0.5pt}lcccc@{\hskip 0.5pt}}
& \multicolumn{2}{c}{NEGRA}
& \multicolumn{2}{c}{TIGER}
\\
Parser & F1 & Disc. F1 & F1 & Disc. F1 \\
\hline
\small{\textit{Predicted POS tags}}\\
\ \ \citet{reduction}    & 77.0 & &  77.3 & \\
\ \ \citet{Cranenburgh2016}, $\leq$40  & 74.8 &   & & \\
\ \ \citet{Versley2016}     & & &  79.5 & \\
\ \ \citet{stanojevic2017}     & & &   77.0 & \\
\ \ \citet{coavoux2017}     & & &   79.3 & \\
\ \ \citet{gebhardt2018}     & & &   75.1 & \\
\ \ \citet{coavoux2019a}     & 83.2 & 54.6 &   82.7 & 55.9 \\
\ \ \citet{coavoux2019b}     & 83.2 & 56.3 &   82.5 & 55.9 \\
\ \ \textbf{This work} & \textbf{85.4} & \textbf{58.8} &  \textbf{85.3} & \textbf{59.1} \\
\hline
\small{\textit{Gold POS tags}}\\ 
\ \ \citet{maier2015} & 77.0 & 19.8 & 74.7 & 18.8 \\
\ \ \citet{reduction}    & 80.5 & & 80.6 & \\
\ \ \citet{maier2016}    & & & 76.5 & \\
\ \ \citet{corro2017}  &  & & 81.6 & \\
\ \ \citet{stanojevic2017}     & 82.9 & & 81.6 & \\
\ \ \citet{coavoux2017}     & 82.2 & 50.0 & 81.6 & 49.2 \\
\ \ \textbf{This work} & \textbf{86.1} & \textbf{59.9} & \textbf{86.3} & \textbf{60.7} \\
\hline
\hline
\textbf{This work} &  \textbf{} & & & \textbf{}  \\
\ \ \ \ \ \textbf{w/o POS tags} & 85.7 & 58.6 & 85.7 & 60.4 \\
\ \ \ \ \ \textbf{w/o POS tags, w/o pre-trained word embs.} & 83.7 & 54.7 & 84.6 & 57.9 \\
\hline
\multicolumn{1}{c}{}\\
\end{tabular}
\centering
\setlength{\abovecaptionskip}{4pt}
\caption{Accuracy comparison of state-of-the-art discontinuous constituent parsers on NEGRA and TIGER. 
}
\label{tab:results}
\end{table*}

\section{Related work}
\cite{Hall2008} was the first work that reduced discontinuous constituent trees into a non-projective dependency representation so that the advances in dependency parsing could be easily adapted to build phrase-structure trees. The main weakness of this first attempt was the spine enconding scheme. They encoded the whole dependent spines (compounded by one or more non-terminal symbols) plus the attachment position in the head spine into complex arc labels, leading to a combinatorial explosion and a significantly large number of labels. This considerably harmed final parsing performance. Following the same encoding scheme, \cite{versley2014} introduced an easy-first approach that improved accuracy and speed, but the complex label scheme was still penalizing parsing accuracy.

\cite{reduction} presented a lighter encoding strategy. They proposed to encode spine interactions by just saving the non-terminal node that results from the spine attachment and the position where it is undertaken. Unlike the previous complex scheme, this cannot deal with unary constituents that should be recovered as a post-processing step. In addition, they used a pre-deep-learning parser, \textit{Turboparser} \cite{turboparser}, that not only reported parsing accuracies notably below the current state-of-the-art, but also was not able to perform a competitive labeling due to the still high number of labels. As a solution, they just performed labeling as a separate step after the parsing was done. The resulting approach notably boosted final accuracy.

The latest attempt to deal with discontinuous phrase-structure parsing as a dependency problem was \cite{corro2017}. Based on the work by \cite{carreras08}, that shows how lexicalized spinal TAGs (Tree Adjoining Grammars) can be used to perform continuous constituent parsing, they proposed a variant for discontinuous structures. Their approach works by reducing spinal TAG parsing to a maximum spanning tree problem and using a neural graph-based dependency parser \cite{DozatM17} to solve it. To simplify the process, they also perform labeling after dependency trees are built.

In our approach, we follow the simplest grammar-agnostic encoding scheme introduced by \cite{reduction} and develop a state-of-the-art neural network architecture that is able to produce an output in a single step by jointly learning to create and label augmented dependency arcs. We consider that this one-stage approach is crucial since arcs and labels are strongly interdependent and they jointly encode the constituent structure.

\section{Conclusions}
We present a novel neural architecture that is able to produce the most accurate discontinuous constituent representations reported so far on the two main benchmarks: NEGRA and TIGER treebanks. We rely on Pointer Networks \cite{Vinyals15} and a biaffine classifier \cite{DozatM17} to efficiently generate discontinuous phrase-structure trees. 
The proposed grammar- and data-structure-agnostic neural model is able to accurately process the large amount of labels generated by the formalism introduced by \cite{reduction} and, unlike the original work, provides an output in a one-stage procedure. In addition, our approach is orthogonal to other techniques such as re-ranking or semi-supervision that can certainly boost final accuracy.

\section{Acknowledgments}
This work has received funding from the European
Research Council (ERC), under the European
Union's Horizon 2020 research and innovation
programme (FASTPARSE, grant agreement No
714150), from the ANSWER-ASAP project (TIN2017-85160-C2-1-R) from MINECO, and from Xunta de Galicia (ED431B 2017/01).

\fontsize{9.0pt}{10.0pt}
\selectfont

\bibliography{main,twoplanaracl,bibliography}
\bibliographystyle{aaai}

\end{document}